\documentclass[11pt]{article}

\usepackage{emnlp2021}

\usepackage{times}
\usepackage{latexsym}

\usepackage[T1]{fontenc}

\usepackage[utf8]{inputenc}

\usepackage{microtype}

\usepackage{hyperref}       
\usepackage{booktabs}       
\usepackage{amsfonts}       
\usepackage{nicefrac}       

\usepackage{xfrac}
\usepackage{xcolor}

\usepackage[rightcaption]{sidecap} 
\sidecaptionvpos{figure}{t}

\usepackage{url} 
\PassOptionsToPackage{hyphens}{url}\usepackage{hyperref} 

\usepackage{graphicx} 
\graphicspath{ {images/} } 

\usepackage{comment} 

\usepackage{amssymb} 
\usepackage{amsmath} 

\usepackage{booktabs} 

\usepackage{paralist} 

\usepackage{CJKutf8} 

\usepackage{subfigure} 

\usepackage{wrapfig} 

\usepackage{float} 

\newcommand{\ie}{\textit{i}.\textit{e}. }
\newcommand{\eg}{\textit{e}.\textit{g}. }

\definecolor{red}{rgb}{1, 0.3, 0.3}

\newcommand{\abs}[1]{\left\lvert #1 \right\rvert}

\title{Scalable Font Reconstruction with Dual Latent Manifolds}

\author{Nikita Srivatsan \\
  Language Technologies Institute \\ Carnegie Mellon University \\
  \texttt{nsrivats@cmu.edu} \\\And
  Si Wu \\
  Khoury College of Computer Science \\
  Northeastern University \\
  \texttt{siwu@ccs.neu.edu} \\\AND
  Jonathan T. Barron \\
  Google Research \\
  \texttt{barron@google.com} \\\And
  Taylor Berg-Kirkpatrick \\
  Computer Science and Engineering \\ University of California, San Diego \\
  \texttt{tberg@eng.ucsd.edu}}

\begin{document}

\maketitle

\begin{abstract}
We propose a deep generative model that performs typography analysis and font reconstruction by learning disentangled manifolds of \emph{both} font style and character shape.
Our approach enables us to massively scale up the number of character types we can effectively model compared to previous methods.
Specifically, we infer separate latent variables representing character and font via a pair of inference networks which take as input sets of glyphs that either all share a character type, or belong to the same font.
This design allows our model to generalize to characters that were not observed during training time, an important task in light of the relative sparsity of most fonts.
We also put forward a new loss, adapted from prior work that measures likelihood using an adaptive distribution in a projected space, resulting in more natural images without requiring a discriminator.
We evaluate on the task of font reconstruction over various datasets representing character types of many languages, and compare favorably to modern style transfer systems according to both automatic and manually-evaluated metrics.
\end{abstract}


\section{Introduction}
\label{sec:intro}

\begin{figure}
\centering
\includegraphics[trim=25 40 40 20 , width=0.82\columnwidth]{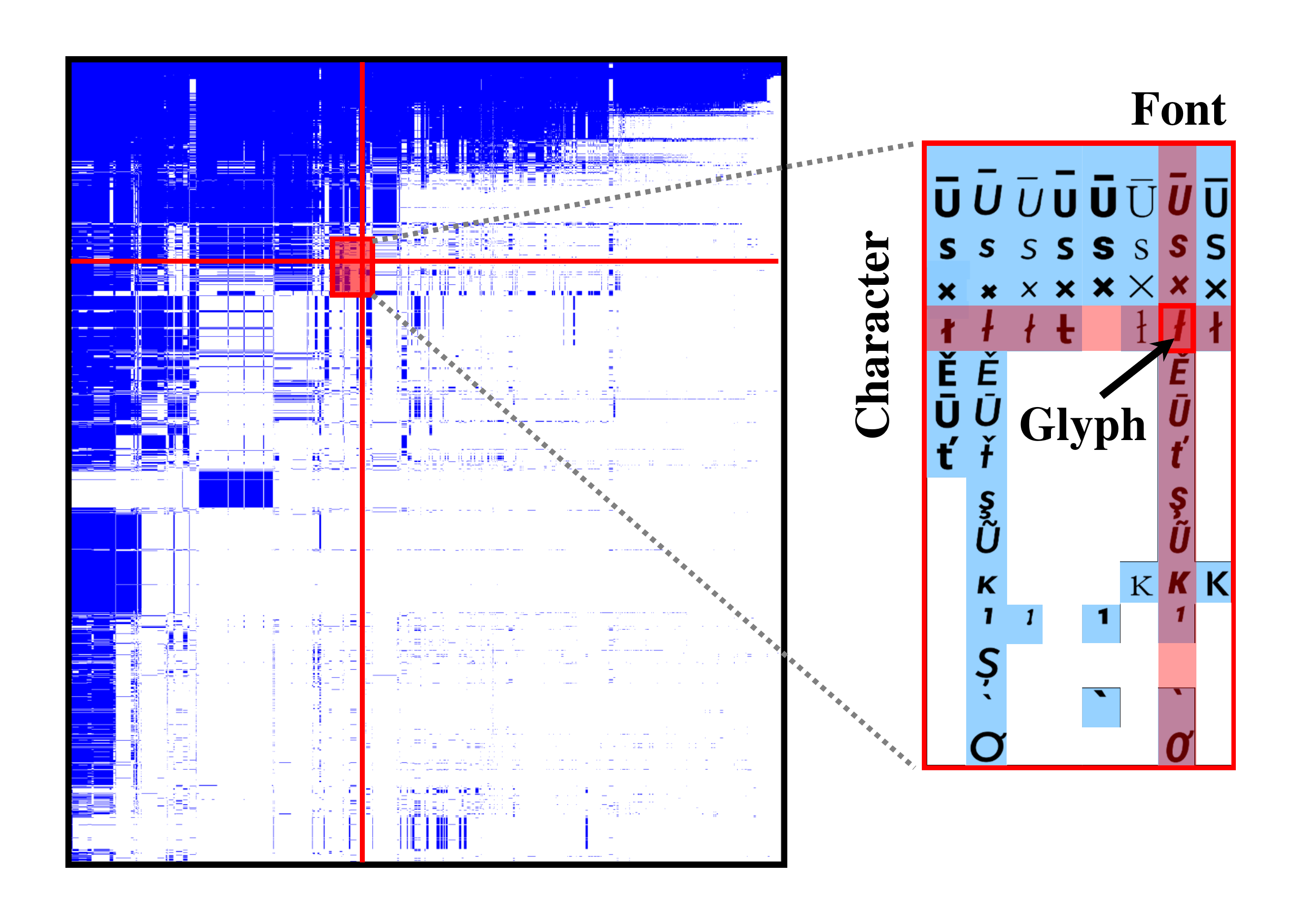}
\caption{Supported glyphs in Google Fonts organized by character type and font. A blue pixel indicates that column's font includes that row's character. Our proposed model allows font reconstruction over this large, sparse character set.}
\vspace{-3mm}
\label{fig:histogram}
\end{figure}

The majority of written natural language comes to us in the form of glyphs, visual representations of characters generally rendered in a font with a contextually appropriate style.
In order to be legible a glyph must be recognizable as its corresponding character from its underlying shape, but it must also be stylistically consistent with the other glyphs in that font.
While this labor intensive process is typically performed manually by human artists, the number of character types that a font may be expected to support is extremely large, with \texttt{Unicode 13.0.0} including as many as 143,859 character types~\citep{unicode}.
As a result, graphic designers often create glyphs only for a subset of these characters, which tends to be determined by their own cultural context.
This can create an accessibility gap for users seeking to create or read digital content in languages with less widespread orthographies, due to the relative lack of available options.
Figure~\ref{fig:histogram} shows that within the Google Fonts library~\citep{googlefonts} there is a long tail of fonts with a large proportion of missing glyphs.

Font reconstruction is a task that attempts to solve this problem.
The goal is for a model, given a small set of example glyphs from an incomplete font, to generate glyphs for the remaining characters in a consistent style.
Prior work has approached this in various ways, albeit with some limitations.
While some approaches use a variational framework~\cite{deepfonts,svgfonts}, generally their models only treat the font style as a latent variable, and not the character shape.
Such methods can therefore only handle a small, fixed, and an a priori known set of character types.
Other work such as that by by~\citet{EMD,gao2019artistic} use discriminative models which dynamically compute both embeddings, allowing them to generalize to unseen characters.
However these networks typically require a pre-specified number of observations as input, and their lack of a probabilistic prior can lead to learning a brittle manifold on datasets with a large number of infrequently observed characters.

By contrast, our method learns two smooth manifolds over character shape and font style in order to better share parameters across structurally similar characters, letting it scale to a larger set and more effectively generalize to characters never seen during training.
Our model treats font reconstruction as a matrix factorization problem, where we view our corpus as a matrix with rows corresponding to character type, and columns corresponding to fonts.
Each row and column is assigned a latent variable that determines its structure or style respectively.
A decoder network consisting of transposed convolutional layers parameterizes the model's distribution on each cell in that matrix, \ie an image of a glyph, conditioned on the corresponding row and column embeddings.
This approach can be thought of as a generalization of~\citet{deepfonts}, who used a similar factorization framework, but with only one manifold over font style.

In addition to model structure, the loss function is also important in font reconstruction as pixel independent losses like $L_2$ tend to produce blurry output, reflecting an averaged expectation instead of something realistic.
Some have used generative adversarial networks (GANs) to mitigate this~\cite{GAN}, but these can suffer from missing modes and collapse issues.
We instead introduce a novel adaptive loss to font reconstruction that operates on a wavelet image representation, while still permitting a well formed likelihood.

Specifically, in this paper we make the following contributions:
(1) Propose the ``Dual Manifold'' model which treats both style and structure representations as latent variables
(2) Propose a new adaptive loss function for synthesizing glyphs, and demonstrate its improvements over more common losses
(3) Put forward two datasets that emphasize few-shot reconstruction, along with a preprocessing technique to remove near-duplicate fonts resulting in more challenging train/test splits.

We evaluate on the task of few-shot font reconstruction, reporting the structural similarity (SSIM) -- a popular metric for image synthesis better correlated with human judgement than $L_2$~\cite{ssim} -- between reconstructions and a gold reference.
These experiments are further split into \textit{known} characters, which the model observed in at least one font at train time, and \textit{unknown} characters, which can be thought of as a few-shot task.
In addition we also perform human evaluation using Amazon Mechanical Turk.
Our approach outperforms various baselines including nearest neighbor, the single manifold approach we build on \cite{deepfonts}, and the previously mentioned discriminative model \cite{EMD}.

\begin{figure*}
    \centering
    \begin{minipage}[c]{0.49\textwidth}
    \includegraphics[trim= 35 20 20 20, width=1\linewidth]{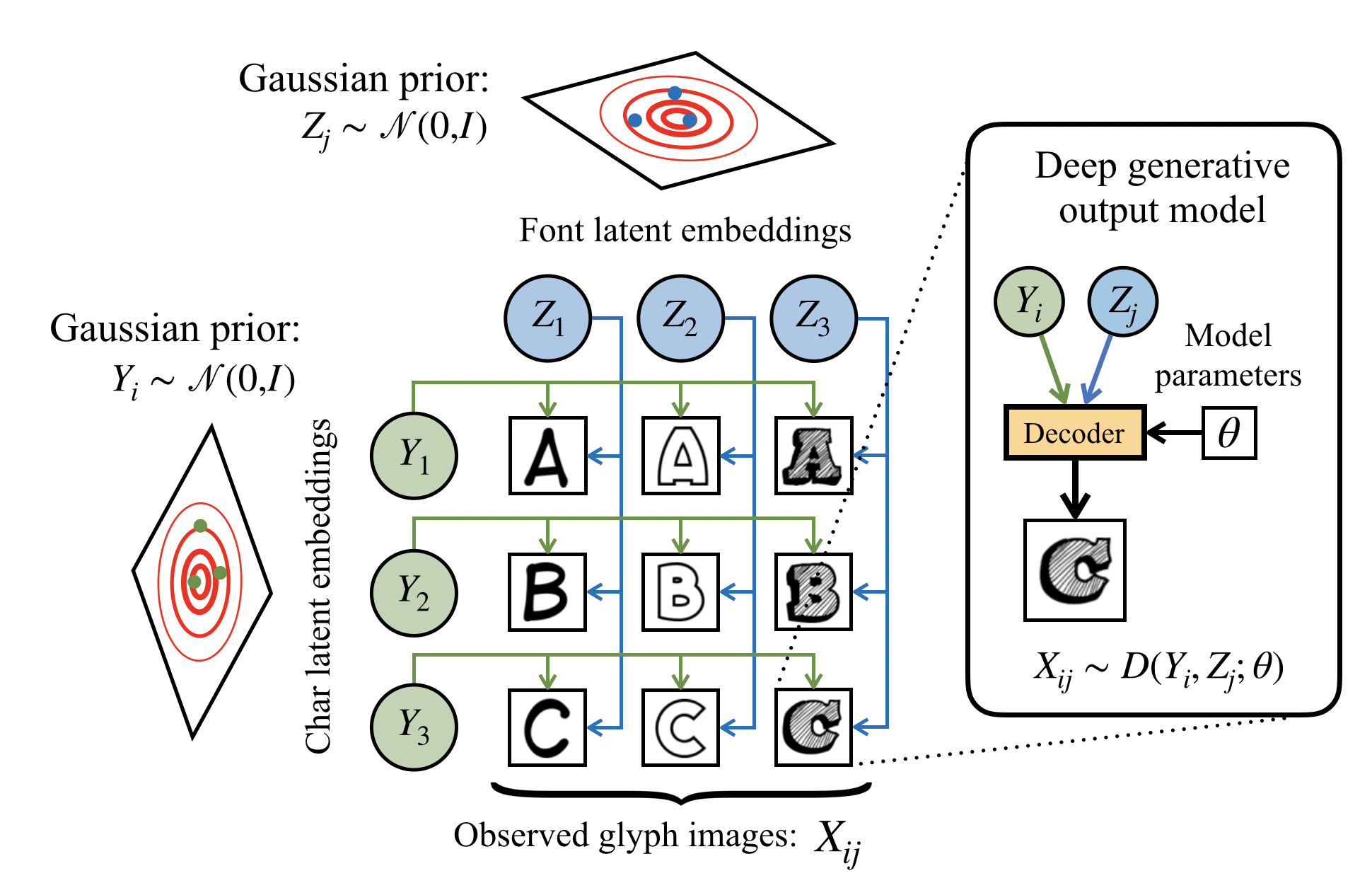}
    \end{minipage} \hfill
    \begin{minipage}[c]{0.5\textwidth}
    \includegraphics[trim= 0 0 0 0, width=1\linewidth]{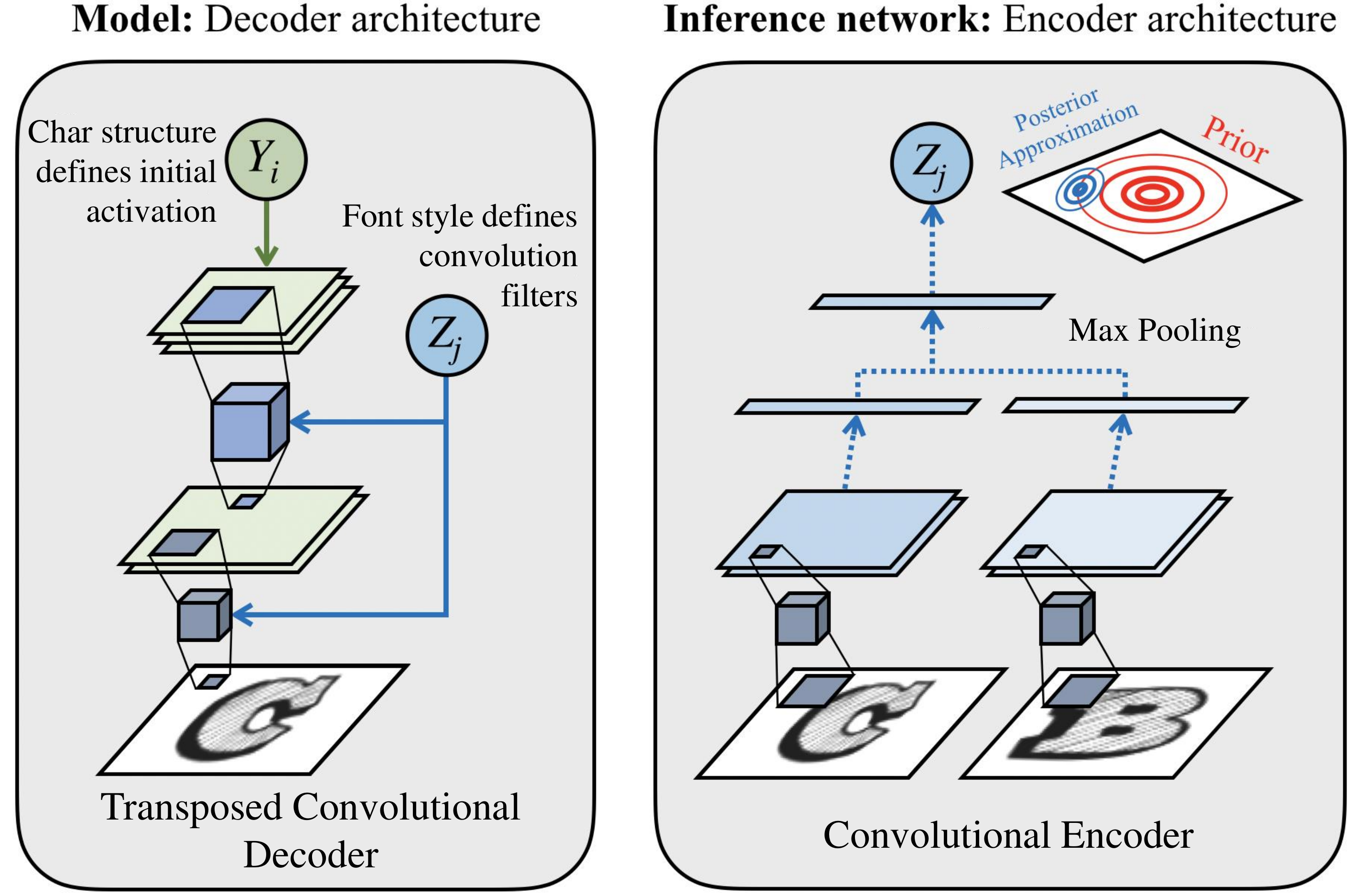}
    \end{minipage} \hfill
    \caption{Overview of our generative factorization model, and important architecture details. Glyphs in the same row share a latent variable $Y_i$ representing character shape, and those in the same column share $Z_j$ representing font style. These variables are inferred by a network that takes in an entire row or column. Our decoder combines these representations to output a distribution on the glyph image.}
    \vspace{-3mm}
    \label{fig:model}
\end{figure*}

\section{Related Work}

A variety of style transfer work has focused specifically on font style, and therefore, font reconstruction.
Some approaches have sought to model the style aspect as a transformation on an underlying topological or stroke-based representation which must be learned for each character \cite{campbell2014learning,phan2015flexyfont,suveeranont2010example}.
However this requires characters to have consistent topologies across fonts.
Other work has learned a font skeleton manifold using Gaussian Process Latent Variable Models~\cite{easyfont}.
One of the more philosophically similar approaches to ours is the bilinear factorization model of Freeman and Tenenbaum~\cite{freeman1997learning,tenenbaum2000separating} which also learns vector representations for each font and character type, albeit in a non-probabilistic and linear manner.
Some more recent research has treated font reconstruction as a discriminative task, using modern neural architectures and techniques from the style transfer literature~\cite{EMD,zhang2020unified,GAN,gao2019artistic}.
Furthermore the concept of learning manifolds for Chinese characters based on shared structure has also been studied~\cite{cw2vec}, albeit with different downstream goals.
\citet{svgfonts} used VAEs which do not observe font alignment across glyphs, but condition on the character type (this work also primarily focuses on generating vector instead of pixel representations).
Finally, more general-purpose style transfer methods for images are well explored~\cite{gatys2015neural,yang2019tet,johnson2016perceptual,wang2016generative,kazemi2019style,chen2017stylebank,ulyanov2016texture}, although these largely lack inductive biases specially suited to typography.

\section{Dual Manifold Model}

\citet{deepfonts} is the most similar prior work, as it also builds from a matrix factorization framework, and learns a latent manifold over font embeddings.
Our model generalizes theirs by learning a second manifold over character shape, letting us massively scale up the number of characters that can be modeled.
In Section~\ref{sec:loss} we also describe our novel loss.

Figure~\ref{fig:model} depicts our model's generative process.
For a corpus consisting of $J$ fonts, each defined over up to $I$ character types, we characterize each particular glyph image as a combination of properties relating to the style of that particular font and to the shape of that character.
Our model effectively factorizes the data by assigning a vector representation to every row and column which correspond to character and font respectively.
Therefore, our approach works by leveraging the fact that all glyphs of the same character type (\ie an entire row in our data) share the same underlying structural shape, and all glyphs within the same font (\ie an entire column) share the same stylistic properties.
By forming separate representations over each of these two axes of variation, we can reconstruct missing glyphs in our data by separately inferring the relevant row and column variables, and then pushing new combinations of those inferred variables through our generative process.
This can be thought of as a form of matrix completion, where unobserved entries correspond to characters not supported by particular fonts.

\newcommand{\x}{\mathbf{x}}
\newcommand{\y}{\mathbf{y}}
\newcommand{\z}{\mathbf{z}}
\newcommand{\weights}{\boldsymbol \theta}

Given a corpus $X$ consisting of $I$ characters across $J$ fonts, we assign to each observed glyph $X_{ij}$ a pair of latent variables which model the properties of that glyph's character type and font style.
Specifically we define these as $Y_i \in \mathbb{R}^{k}$ and $Z_j \in \mathbb{R}^{k}$, which we draw from a standard Gaussian prior $\mathcal{N}(0, I_{k})$, with $Y$ modeling the shape of the character (e.g. a \texttt{q} or \texttt{<}), and $Z$ modeling the properties of the font (e.g. \textit{Times New Roman} or \textit{Roboto Light Italic}).
Given a particular $Y_i$ and $Z_j$, we combine them via a neural decoder to obtain a distribution $p(X_{ij} | Y_i, Z_j ; \theta)$ which scores the corresponding glyph image $X_{ij}$. This yields the following likelihood function:
\begin{equation*}
\resizebox{\columnwidth}{!}{$
    p(X, Y, Z ; \theta) = \prod_{I} p(Y_i) \prod_{J} p(Z_j) \prod_{I,J} p(X_{ij} | Y_i, Z_j ; \theta)
$}
\end{equation*}
Both $Y$ and $Z$ are unobserved, and we must therefore infer both to train our model and produce reconstructions at test time.
Note that by contrast,~\citet{deepfonts} represents characters as fixed parameters, and must only perform inference over font representations.
We use a pair of encoder networks to perform amortized inference, as depicted in Figure~\ref{fig:encoder}.

\subsection{Decoder Architecture}

The basic structure of our decoder is largely identical to the popular U-Net architecture~\cite{unet} which has seen much success on image generation tasks with its coarse-to-fine layout of transposed convolutional layers.
However, we make a few key modifications (depicted in Figure~\ref{fig:model}) in order to imbue our decoder with stronger inductive bias for this particular task.
Following~\citet{deepfonts}, instead of directly parameterizing the transposed convolutional layers that appear within each block of the network, we allow the weights of each layer to be the output of an MLP that takes as its input the font variable $Z_j$.
This is effectively a form of HyperNetwork~\cite{hypernetworks}, a framework in which one network is used to produce the weights of another.
In this way, the parameters of the transposed convolutional layers are dynamically chosen based on the font variable.
By contrast, $Y_i$ is the input fed in at the top of the decoder, to which these filters are applied.
The purpose of this asymmetry is to encourage $Z_j$ to learn properties relating to finer stylistic information, while $Y_i$ learns more spatial information about the characters.
In another manner of speaking, $Y_i$ should learn ``what'' to write, and $Z_j$ should learn ``how'' to write it.

\section{Adaptive Wavelet Loss}
\label{sec:loss}

\newcommand{\cdf}{\operatorname{\psi}}

Our decoder architecture outputs a grid of values, but an important decision is what distribution (and therefore loss) these should finally parameterize to score actual pixels.
Traditional approaches using variational autoencoders have modeled each pixel as an independent Normal distribution, which results in the model minimizing the $L_2$ loss between its output and gold.
This however leads to oversmoothed images, as it treats adjacent pixels as independent despite their strong correlations~\cite{bell1997edges}, and fails to account for the heavy-tailed distribution of oriented edges in natural images~\cite{Field87}.
As a result $L_2$ penalizes the model for generating images that are realistic but slightly transposed or otherwise not perfectly aligned with gold, which encourages models to produce fuzzy edges in order to be closer on average.
GANs are often employed to force sharper output~\cite{GAN,gao2019artistic}, but following recent work we instead use a projected loss for a similar effect.

At a high level, our approach will first project images to a feature space, and let the model's output parameterize a distribution on this projection.
If that projection is invertible and volume-preserving, this is equivalent to directly parameterizing a distribution on pixels, but allows for more expressivity~\cite{rezende2015variational,dinh2014nice,dinh2016density}.
Ideally, such a loss requires a distribution expressive enough to capture the variable frequency characteristics of natural images, and a representation of the image that explicitly reasons about spatially-localized edges.

A good example of this technique is that of~\citet{deepfonts}, which modeled images by placing a Cauchy distribution on a 2D Discrete Cosine Transform (DCT) representation of glyphs.
Though this is an improvement over the default choice of placing a Normal distribution on individual pixels as it both decorrelates pixels and is tolerant of outliers, this approach is limited in its expressiveness and its ability to model spatially localized edges:
Cauchy distributions are excessively heavy-tailed and so have difficulty modeling inliers, and since DCT is a global representation it does not allow the model to reason about \emph{where} image gradient content is located.

\begin{figure}[H]
\centering
\includegraphics[trim=0 50 0 0, width=0.7\columnwidth]{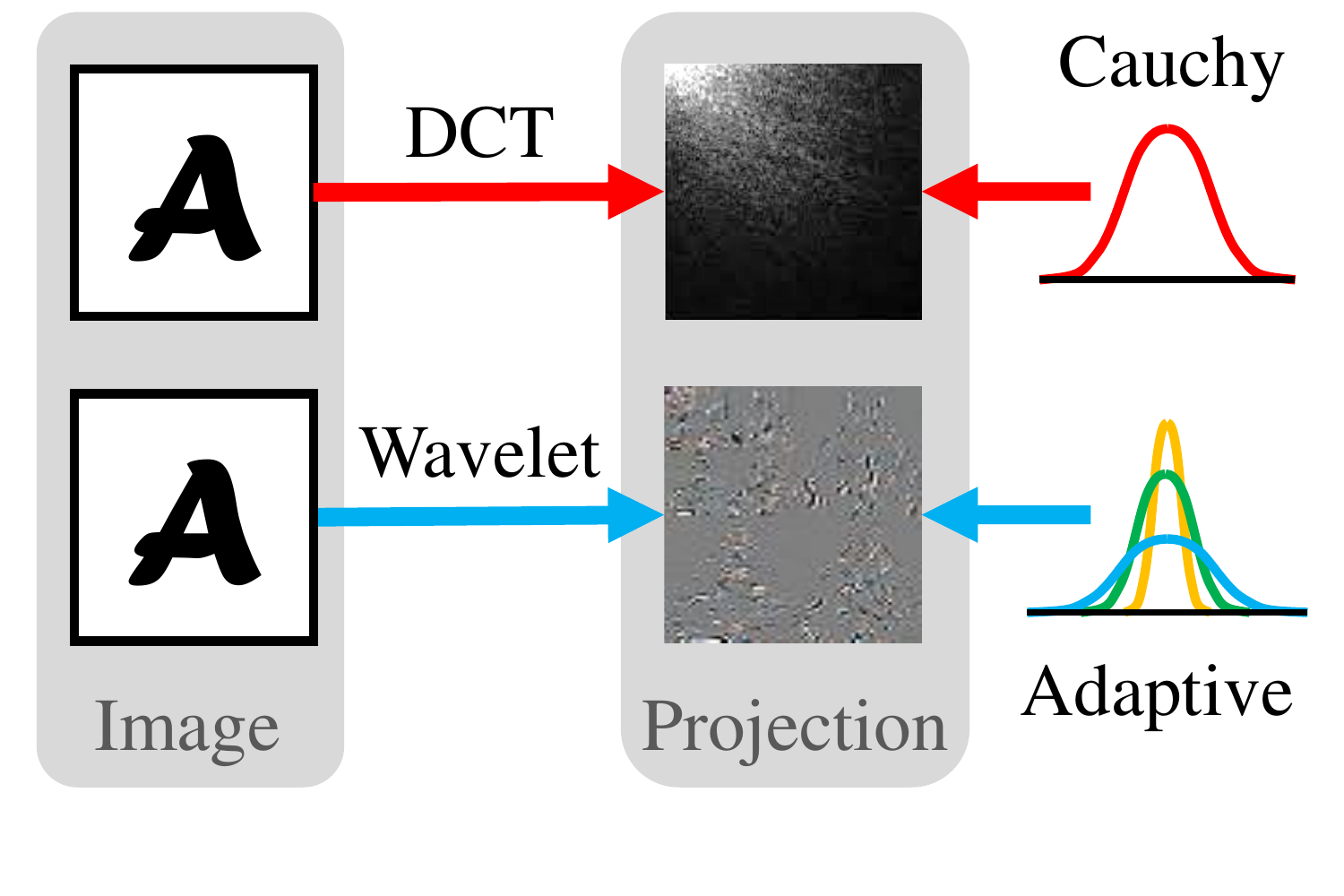}
\end{figure}

We extend this approach in two ways (as depicted above):
(1) by using a wavelet image representation instead of DCT,
and (2) by using a distribution with an adaptive shape instead of a Cauchy.

\begin{SCfigure*}
\centering
\includegraphics[width=1.53\linewidth]{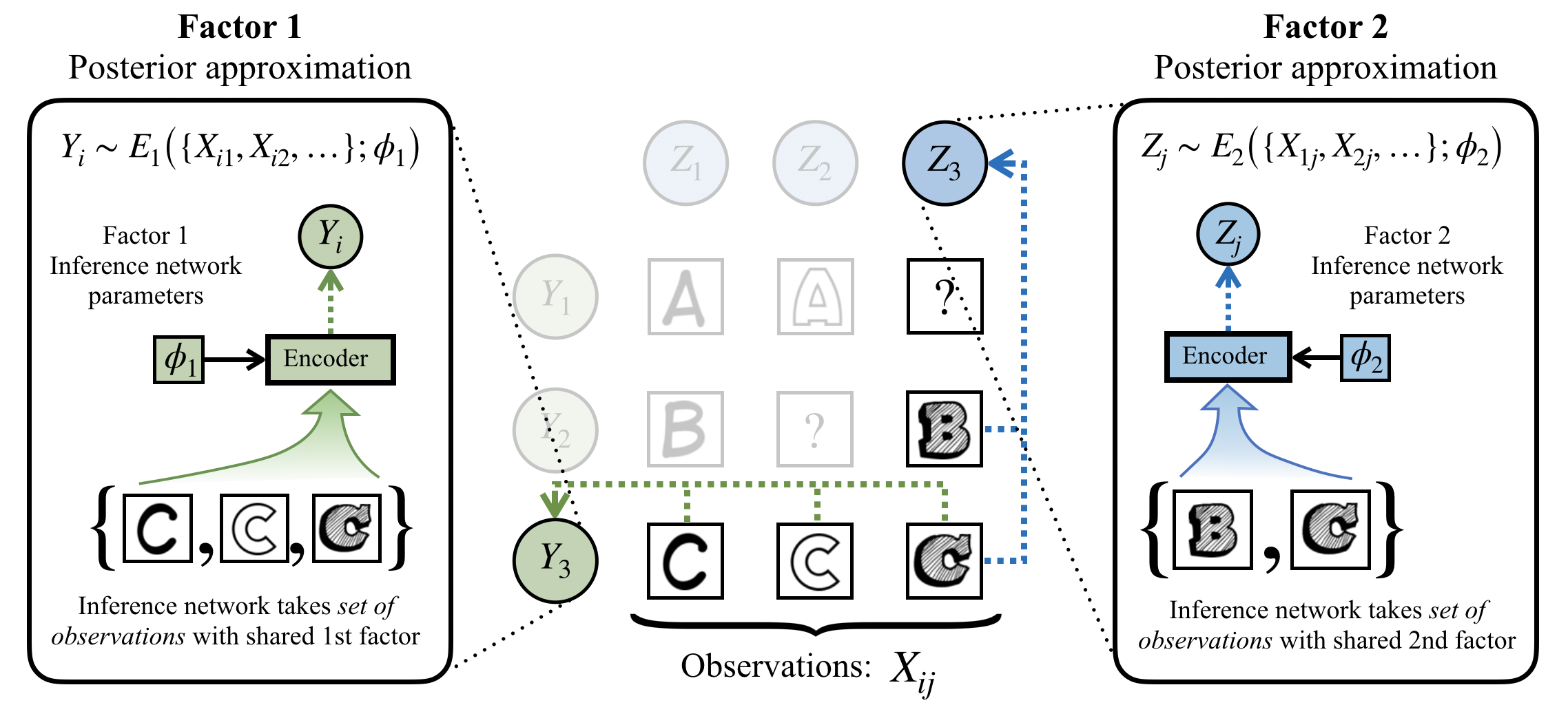}
\caption{Overview of the inference procedure. First the character encoder infers a representation of structure over each row, and then the font encoder infers a representation of style conditioned on a (perhaps partially observed) column and the character embeddings.}
\vspace{-3mm}
\label{fig:encoder}
\end{SCfigure*}

\textbf{Representation} 
We opt for a wavelet representation, as unlike DCT it jointly encodes the frequency \textit{and spatial location} of an image feature.
As might be expected, an image representation in which location is directly encoded is helpful in our task; a stroke has a fundamentally different meaning at the top of a character than at the bottom.
\citet{BarronCVPR2019} quantitatively demonstrated the advantages of specifically the Cohen-Daubechies-Feauveau (CDF) 9/7 wavelet decomposition~\cite{cohen1992biorthogonal} for training likelihood-based models of natural images. 
Based on their findings that CDF 9/7 in front of an adaptive loss achieves better performance than DCT in front of a Cauchy (the setup of~\citet{deepfonts}), we expect similar performance benefits in the context of our own model, and our ablations in Table~\ref{tab:ssim} (Right) support this belief empirically.

\textbf{Distribution} An adaptive distribution lets the model select between using leptokurtotic (Cauchy-like) distributions that are well suited to the high-frequency image edges found at the finer levels of the wavelet decomposition, or more platykurtic (Normal-like) distributions that are better suited to low-frequency DC-like average image intensities found at the coarsest levels of the wavelet decomposition.
Specifically, we use the probability distribution of~\citet{BarronCVPR2019}:
\begin{equation*}
\resizebox{\columnwidth}{!}{$
f(x | \mu, \sigma, \alpha) = \frac{\exp \left( -{ \frac{\abs{\alpha - 2}}{\alpha} } \left( \left( \frac{\left( x - \mu \right)^2}{\sigma^2 \abs{\alpha - 2} } + 1 \right)^{\sfrac{\alpha}{2}} - 1 \right) \right)}{\sigma Z(\alpha)}
$}
\end{equation*}
where $Z(\alpha)$ is the distribution's partition function, and $\alpha$ determines the distribution's shape.
As $\alpha \rightarrow 0$  the distribution approaches a Cauchy distribution, as $\alpha \rightarrow 2$ the distribution approaches a Normal distribution.

Taken together, these yield a conditional likelihood function parameterized by the decoder of our variational model, which we now describe.
Given an image $X_{i,j}$, we first project it using the CDF 9/7 wavelet decomposition -- which we denote as $\cdf(X_{i,j})$.
Because this decomposition is a biorthogonal volume-preserving transformation, it can be applied before the likelihood computation.
It further serves as a whitening transformation, avoiding the need to learn a covariance matrix for $X_{i,j}$.

Our decoder outputs a grid of parameters $\hat{X}_{i,j}$, the projection of which serves as the mean $\mu$ of our adaptive distribution for scoring $\cdf(X_{i,j})$.
For the other distribution parameters $\sigma$ and $\alpha$, rather than using fixed settings we construct a set of latent variables for both: we allow each wavelet coefficient to have its own vector of latent shape parameter $\boldsymbol \ell^{\alpha}$ and scale parameter $\boldsymbol \ell^{\sigma}$, where the non-latent shape and scale are parameterized as scaled and shifted sigmoids and softplus of those latent values:
\begin{equation*}
\resizebox{\columnwidth}{!}{$
\alpha_k = \frac{2}{1 + \exp\left(\ell^{\alpha}_k\right)} ,
\sigma_k = \frac{\log\left(1+\exp\left(\ell^{\alpha}_k\right) \right)}{\log(2)} + \epsilon
$}
\end{equation*}

We initialize $\boldsymbol\ell^{\alpha} = \boldsymbol\ell^{\sigma} = \vec{0}$, thereby initializing $\boldsymbol\alpha = \boldsymbol\sigma = \vec{1}$.
These latent variables $(\boldsymbol\ell^{\alpha}, \boldsymbol\ell^{\sigma})$ are optimized during training using gradient descent along with all other model parameters $\theta$, which allows the model to adapt the shape and scale of each wavelet coefficient's distribution during training.
Overall, this yields the following likelihood function:
\begin{equation*}
\resizebox{\columnwidth}{!}{$
    p(X_{i,j} | Y_i, Z_j ; \theta) = 
     \prod_k f(\cdf(X_{i,j})_k | \cdf(\hat{X}_{i,j})_k,
    \sigma_k, \alpha_k)
$}
\end{equation*}

\section{Learning and Inference}

We now describe our approach to training this model.
This process mirrors that of previous variational work, although since we are learning a dual manifold, our model will require two separate inference networks.
The projected loss we add (Section~\ref{sec:loss}) will not fundamentally affect the learning process, but does change how the reconstruction term is computed.

As our model is generative, we wish to maximize the log likelihood of the training data with respect to the model parameters, which requires summing out the unobserved variables $Y$ and $Z$.
However, this integral is intractable and does not permit a closed form solution.
We therefore resort to optimizing a variational approximation, a strategy which has seen success in similar settings~\cite{vae,deepfonts}.
Rather than directly optimize the likelihood (which we cannot compute the gradient of), we maximize a lower bound on it known as the Evidence Lower Bound (ELBO).
We compute the ELBO via a function $q(Y, Z | X) = q(Y | X) * q(Z | Y,X)$ which approximates the posterior $p(Z, Y | X)$ of the distribution defined by our decoder network.
\begin{equation*}
\resizebox{\columnwidth}{!}{$
    \mathrm{ELBO} = \mathbb{E}_q [ \log p(X | Z, Y) ] - \mathbb{KL} ( q(Z, Y | X) || p(Z)p(Y))
$}
\end{equation*}

We define $q(Y | X)$ and $q(Z | Y, X)$ via a pair of encoder networks which operate over one row or column of the matrix respectively.
An encoder passes each glyph in that row or column through a series of convolutional layers, and then max pools the output features across all glyphs, ensuring it can handle a variable number of observations (See Figure~\ref{fig:encoder}).
Note that the method of pooling (\eg min, max, avg), as well as the order in which to infer $Y$ and $Z$ are important choice points that allow for different inductive biases.
The pooled feature representation is then passed through an MLP which outputs parameters $\mu$ and $\Sigma$ to define a Gaussian posterior over $Y_i$ or $Z_j$.
Given these, we compute approximate gradients on the ELBO via the reparameterization trick described by~\citet{vae}.

\section{Experiments}

We evaluate on the task of font reconstruction, in which given a small random subset of glyphs from a held out font, models must reconstruct the remaining ones.
We separately report performance on known (\ie observed at least once during training) and unknown character types.
During training, we mask out a randomly chosen $20\%$ of character \textit{types} to serve as unknowns.
At test time, models observe examples of previously masked characters to infer their representations for reconstruction.
This can be thought of as a few-shot task, where models must generate glyphs for character types they did not observe at train time based on limited test-time examples.

\subsection{Datasets}

Capitals64, the dataset used by~\citet{GAN} and~\citet{deepfonts}, only contains the $26$ English capital letters, with no missing characters, meaning it does not require learning a manifold over character shape.
We instead evaluate on the following datasets to best demonstrate our method's ability to scale to settings with a large number of character types and a high degree of sparsity.

\textbf{Google Fonts} Google Fonts is a dataset of $991$ font families, which is publicly available\footnote{\url{https://github.com/google/fonts}}.
Most fonts in the dataset support standard Latin characters, but many also support special symbols, and characters found in Greek, Cyrillic, Tamil, and several other orthographies.
A visualization of this is shown in Figure~\ref{fig:histogram}.
We restrict our work to the $2000$ most frequently supported character types for simplicity.
After removing near duplicates (described below) we are left with $2017$ fonts in total, split into train, dev, and test in a $3:1:1$ ratio.
The data was split by font family rather than individual fonts, to ensure that there are no fonts in train with a ``sibling'' in test.

\textbf{Chinese Simplified} We scraped a list of the most common 2000 Chinese simplified characters from the internet as well as a dataset that labels each character's radical. Together, we compile a new dataset that consists of the most common 2000 Chinese characters along with their radicals for further analysis on the character embeddings. For each Chinese character, we scraped over 1524 fonts, split similarly to Google Fonts. The total font number shrinks down to 623 after removing near duplicates, which we now discuss.

\textbf{Removing near duplicates} One major issue with font corpora is that most fonts belong to a small handful of modes, within which there is little stylistic diversity.
To ensure that our metrics best measure generalization to novel fonts unlike those seen in train, we preprocess out fonts that are extremely similar to others in the data.
We first perform agglomerative clustering, and then retain only the centroid of each cluster.
The number of clusters is determined by cutting the dendrogram at a height which eliminates most fonts that are to a human largely indistinguishable from their nearest neighbor.

\subsection{Baselines}

\begin{table*}
\centering
    \begin{minipage}[c]{0.58\textwidth}
        \resizebox{\textwidth}{!}{%
        \begin{tabular}{rcccccccc}
        \toprule
        Observations  & 1 & 8 & 16 & 32 & 1 & 8 & 16 & 32  \\
        \midrule
          & \multicolumn{4}{c}{Google Fonts: Known Char} & \multicolumn{4}{c}{Google Fonts: Unknown Char} \\
        \cmidrule(lr){2-5} \cmidrule(lr){6-9}
        NN       & 0.755 & 0.816 & 0.830 & \textbf{0.839}  & - & -  & - & -          \\
        EMD      & 0.706 & 0.702  & 0.539 & 0.597 & 0.698 & 0.695 & 0.534 & 0.595 \\
        Dual Manifold  & \textbf{0.799} & \textbf{0.828} &  \textbf{0.833} &  0.834  & \textbf{0.801} & \textbf{0.826}  &  \textbf{0.829} & \textbf{0.830} \\
        \midrule
          & \multicolumn{4}{c}{Chinese Simplified: Known Char} & \multicolumn{4}{c}{Chinese Simplified: Unknown Char} \\
        \cmidrule(lr){2-5} \cmidrule(lr){6-9}
        NN       & \textbf{0.428} & \textbf{0.488}  & \textbf{0.495} & \textbf{0.499}  & - & -  & - & -          \\
        EMD      & 0.278 & 0.271  & 0.291  & 0.288 & 0.270 & 0.266 & 0.283 & 0.280 \\
        Dual Manifold  & 0.392 & 0.405 & 0.407  & 0.407    & \textbf{0.375} & \textbf{0.387}  &  \textbf{0.390} & \textbf{0.390} \\
        \bottomrule
        \end{tabular}
        }
    \end{minipage} \hfill
    \begin{minipage}[c]{0.41\textwidth}
        \resizebox{\textwidth}{!}{%
        \begin{tabular}{rcccc}
        \toprule
        Observations  & 1 & 8 & 16 & 32  \\
        \midrule
        +Dual, +KL, +Adapt, Min & 0.7728 & 0.8041 & 0.8083 & 0.8088  \\
        \midrule
        \citet{deepfonts}* & 0.713 & 0.702 & 0.701 & 0.698 \\
        -Dual, +KL, +Adapt, Max & 0.704 & 0.703 &  0.701 &  0.703  \\
        \midrule
        +Dual, -KL, +Adapt, Max & 0.785 & 0.817 & 0.821  & 0.823   \\
        \midrule
        +Dual, +KL, -Adapt, Max & 0.795 & 0.823 & 0.828 & 0.829  \\
        +Dual, +KL, +Adapt, Max & \textbf{0.799} & \textbf{0.828} &  \textbf{0.833} &  \textbf{0.834} \\
        \bottomrule
        \end{tabular}
        }
    \end{minipage} \hfill
\caption{(Left) SSIM per glyph by number of observed characters for Google Fonts and Chinese Simplified. (Right) Ablations of our model, showing SSIM results on known characters in Google Fonts.}

\label{tab:ssim}
\end{table*}

We compare our model -- which we refer to as \textsc{Dual Manifold} -- to two baselines and various ablations.
Our primary baseline is \textsc{EMD}~\cite{EMD}, a discriminative encoder-decoder model that does not share embeddings across ``rows'' and ``columns'', but rather computes style and content representations for each glyph given a set of provided examples, and then passes them to a generator which constructs the final image.
This model is useful for comparison as it has a similar computation graph and also learns separate embeddings for font and character shape, but computes its loss directly in pixel space, and lacks a probabilistic prior.

We also compare to a naive nearest neighbor (\textsc{NN}) model, which reconstructs fonts at test time by finding the font in train with the closest $L_2$ distance over the observed characters, and outputs that neighbor's corresponding glyphs for the missing characters.
If the neighbor does not support all missing characters, we pull the remaining from the 2nd nearest neighbor, and so on.
It should be noted that \textsc{NN} cannot reconstruct any character that is not present in train.

Similarly to \textsc{EMD}, the first of our ablations, denoted \textsc{-KL}, does not explicitly model the character and font embeddings as random variables.
This effectively removes the KL divergence from the loss function, resulting in a non-probabilistic autoencoder.
The next, denoted \textsc{-Dual}, is an ablation which treats the character representations as parameters of the model, rather than latent variables which must be inferred.
This is essentially the model of~\citet{deepfonts} with our architecture*.
We also ablate our adaptive wavelet loss against the DCT + Cauchy loss used by~\citet{deepfonts}, denoted with \textsc{-Adapt}.
Finally, we compare performing \textsc{Max} or \textsc{Min} pooling over elements of a row/column within the encoder network.

\subsection{Training Details}

We perform stochastic gradient descent using Adam~\cite{adam}, with a step size of $10^{-4}$.
Batches contain $10$ fonts, each with only $20$ random characters observable to encourage robustness to the number of inputs.
However at test time, the model can infer the character representations $Y$ based on the the entire training set.
We find best results when both character and style representations are $k=256$ dimensional.
See Appendix~\ref{app:arch} for a full description of architecture.
Our model trains on one NVIDIA 2080ti GPU in roughly a week, and is implemented in PyTorch~\cite{pytorch} version \texttt{1.3.1}

\subsection{Metrics}

We measure average SSIM per glyph~\cite{GAN,gao2019artistic}, having scaled pixel intensities to $[0,1]$.
While the details of SSIM are beyond the scope of this paper, it can be thought of as a feature-based metric that does not factor over individual pixels, but rather looks at the matches between higher level features regarding the structure of the image.
SSIM is widely used in image processing tasks since it measures structural similarity instead of raw pixel distance, and has been shown to better correlate with human judgement than $L_2$~\cite{ssim}.
Evaluating models using $L_2$ can reward unrealistic reconstructions that split the difference between many hypotheses as opposed to picking just one (part of the reason we avoid training our model on such a loss).
Over the course of individual training runs, we found it was almost counter correlated with human judgement, with the lowest distance early in training while output was blurry, becoming larger as the model converged.
We do however include these numbers in Appendix~\ref{app:mse}, as they nonetheless support our findings.

We also perform human evaluation using Amazon Mechanical Turk.
For each font in our test set, 5 turkers were shown 8 example glyphs, and a sample of reconstructions for the remaining characters by \textsc{Dual Manifold} and \textsc{EMD}.
Turkers were also shown examples of each character in a neutral style.
They were asked to select which if either reconstruction was better, and briefly justify their reasoning.

\section{Results}

\subsection{Quantitative Evaluation}

We list SSIM results in Table~\ref{tab:ssim} for various numbers of observed characters.
Note that \textsc{NN} is not capable of reconstructing character types not observed at training time.
On Google Fonts, our model performs best overall; however on Chinese Simplified, we see \textsc{NN} winning on known characters, as well as a marked drop in SSIM overall.
This could be due to the increased challenge in generating Chinese characters given the relatively higher number of strokes, leading SSIM to prefer the realism of \textsc{NN}, or because fonts in this dataset generally contain most characters, unlike Google Fonts which is much sparser.
Observing more characters taperingly increases similarity, which matches our intuition that this allows for a better understanding of stylistic properties.
Performance drops when evaluating on characters not observed in training.
This makes sense as models may have less support in their manifolds for structural forms they were not trained on, but the drop is small enough to suggest our model is able to infer meaningful representations for novel character types at test time.

We see also that \textsc{EMD} has significant issues at 16 and 32 observed characters (it's worth noting that \textsc{EMD} must be separately trained for each number of observations).
Qualitatively, we find certain fonts for which \textsc{EMD} emits the same output for every character in that font.
We suspect this indicates overfitting leading to broken style representations for some novel fonts when given more observations than its default of only 10.

\begin{SCfigure}[5]
    \centering
    \includegraphics[width=0.5\columnwidth]{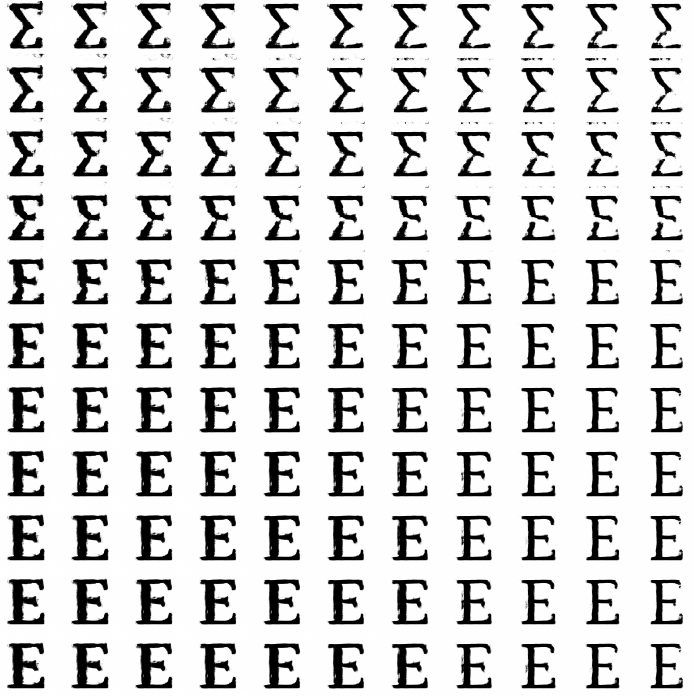}
    \caption{Generated glyphs for interpolating between both character type (horizontal axis) and font style (vertical axis) simultaneously.}
    \label{fig:matrix}
    \vspace{-3mm}
\end{SCfigure}

Within our ablations, we find that using a dual latent manifold, as opposed to treating character embeddings as model parameters, is responsible for the majority of our gain in SSIM over prior work.
The next largest difference comes from using either \textsc{Min} pooling within the autoencoder or \textsc{Max} pooling.
We also see more of a drop in performance from removing the \textsc{KL} divergence, than we do from replacing our adaptive wavelet loss with the DCT + Cauchy loss.

\begin{figure}
    \centering
    \includegraphics[trim=0 20 0 20, width=\columnwidth]{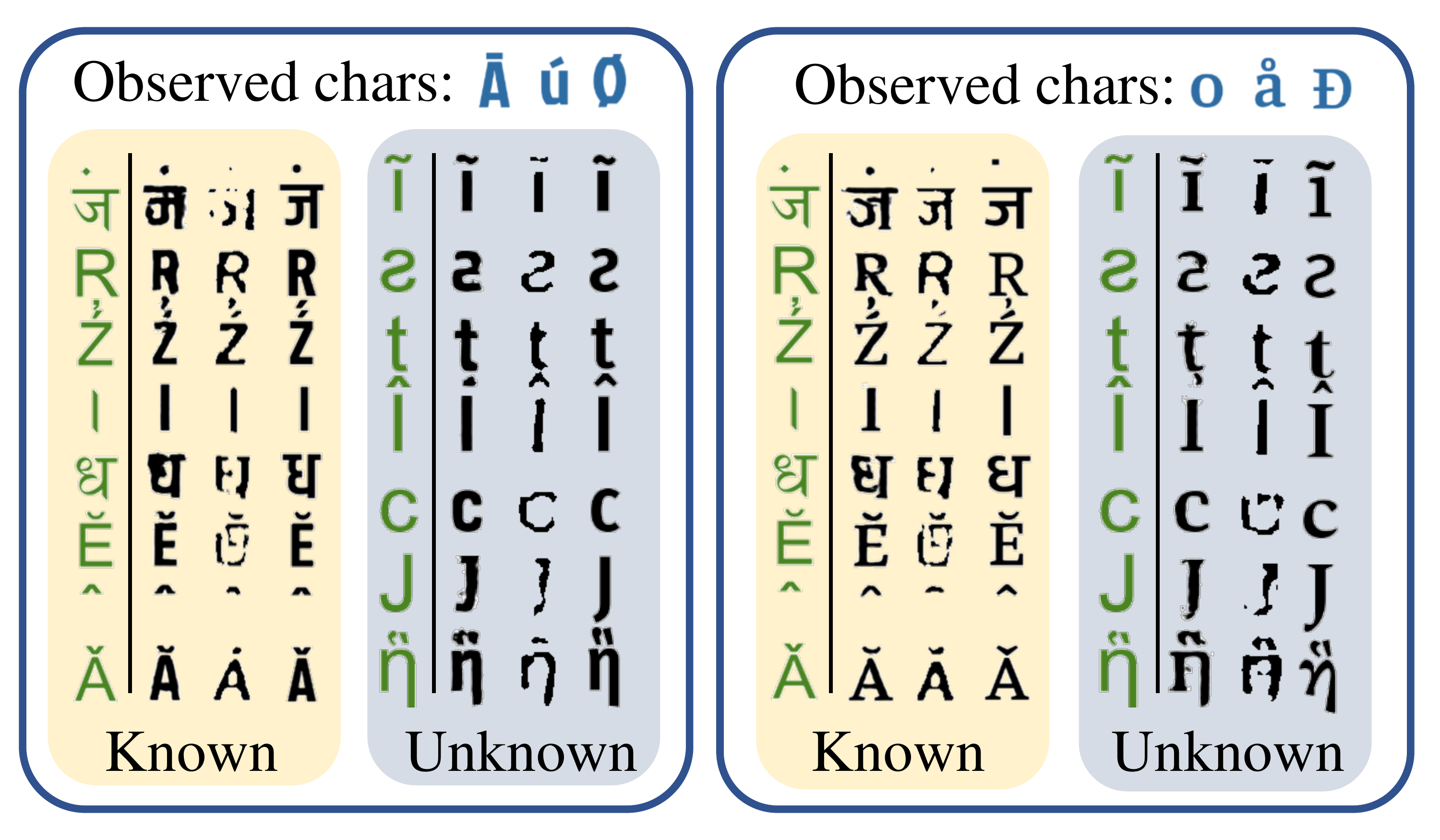}
    \caption{Reconstructions of two fonts from our model, EMD, and NN --- shown in black in that order --- for both known and unknown character types. Green characters show the expected shape in a neutral font, and blue characters are a sample of those observed by the models for either font.}
    \label{fig:qual}
\vspace{-3mm}
\end{figure}

\subsection{Human Evaluation}

\begin{figure*}
  \centering
  \subfigure{\includegraphics[trim=0 30 0 0, scale=0.42]{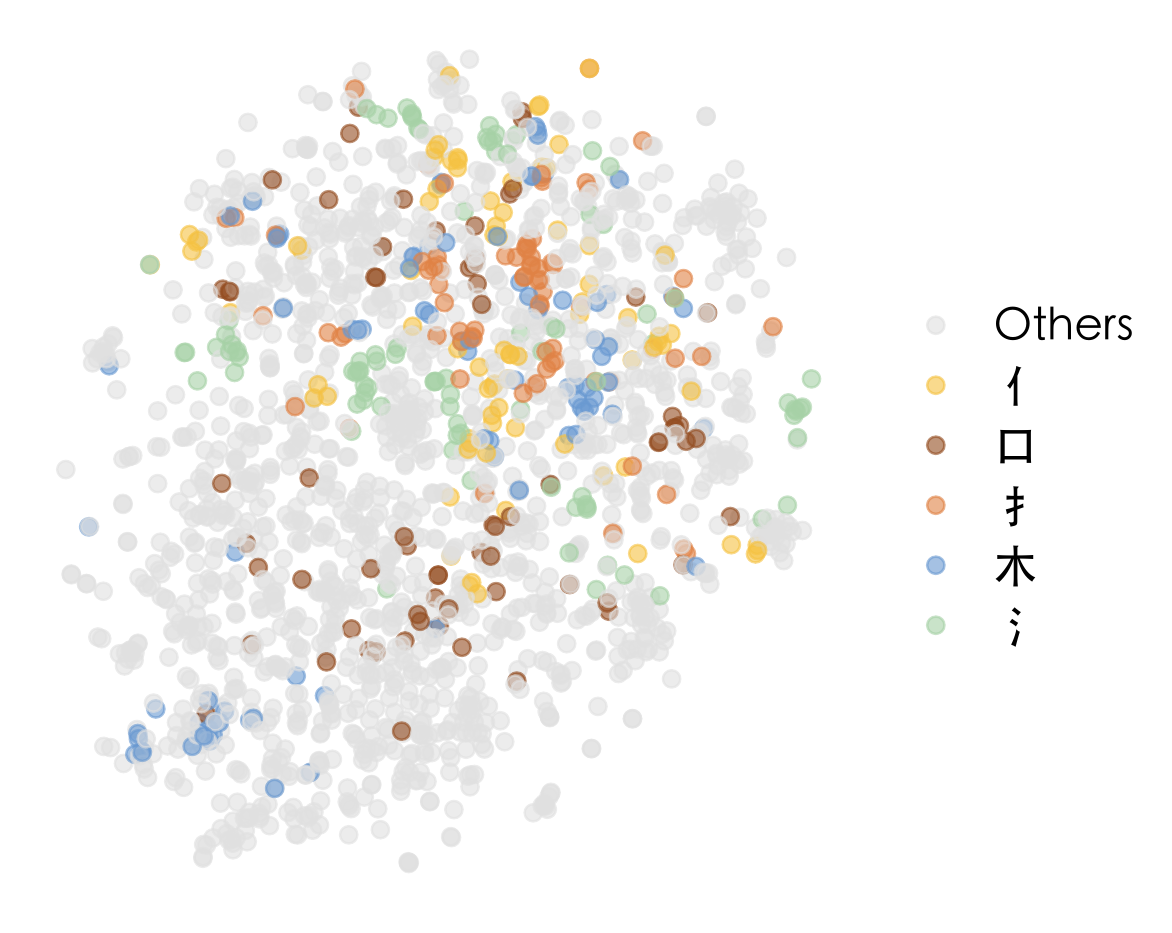}}\quad
  \subfigure{\includegraphics[trim=0 30 0 0, scale=0.42]{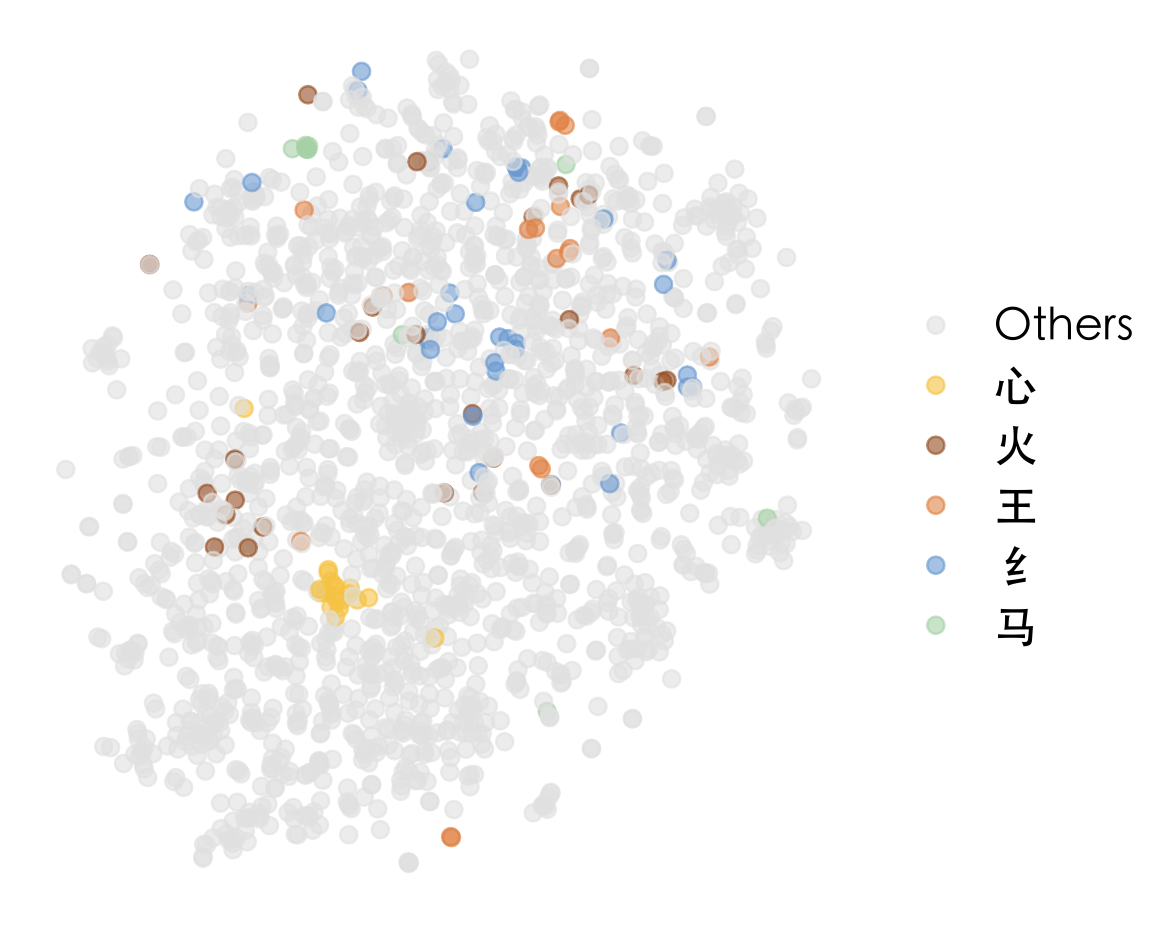}}\quad
  \subfigure{\includegraphics[trim=0 30 0 0, scale=0.42]{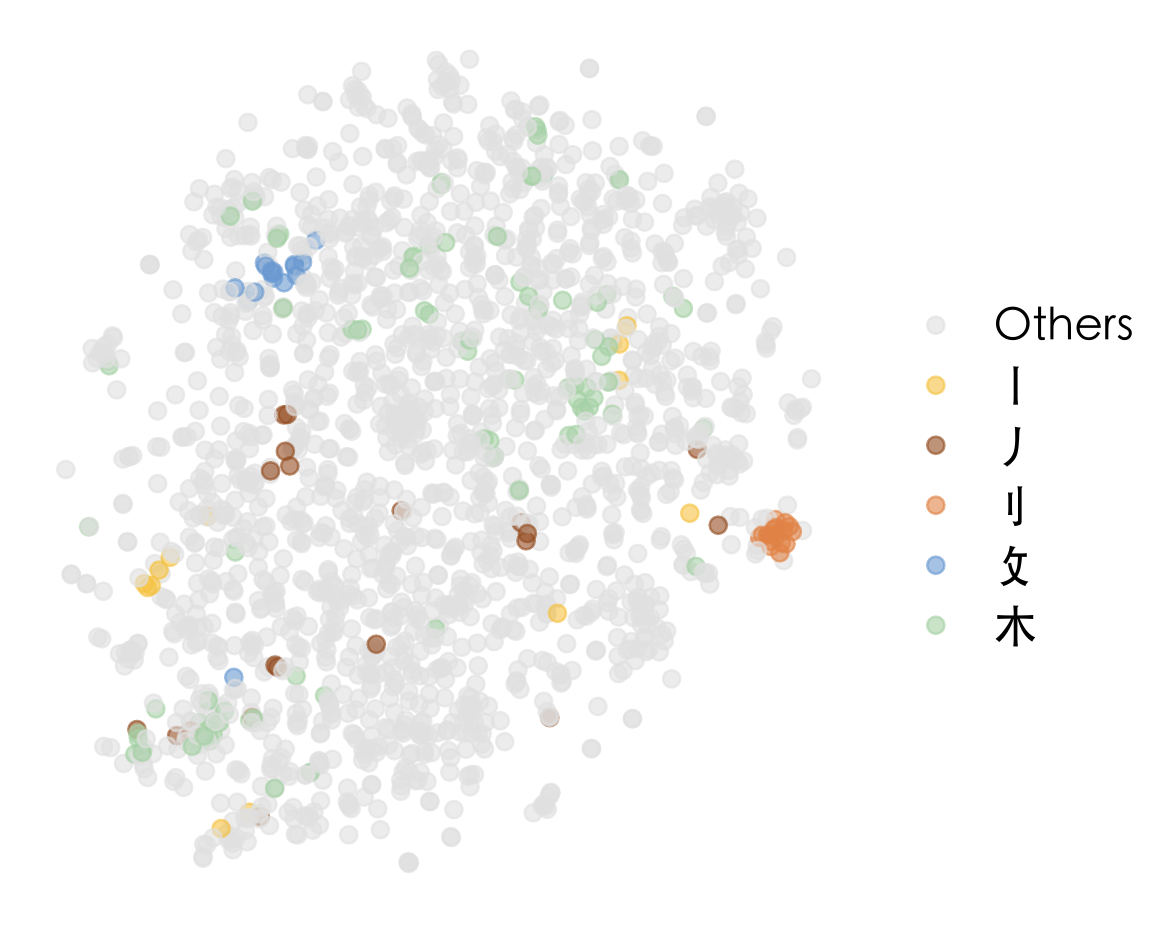}}
\caption{t-SNE plot of Chinese character embeddings from our model for the top 5 radicals (left), and randomly chosen groups of 5 (middle, right).}
\vspace{-2mm}
\label{fig:tsne}
\end{figure*}

In our AMT experiments, we found that for $\textbf{48.2\%}$ of known character reconstructions, turkers preferred our model's output, with $42.0\%$ preferring \textsc{EMD}, and $9.8\%$ finding both equal.
For unknown character reconstructions, $\textbf{50.5\%}$ preferred ours, vs $38.7\%$ for \textsc{EMD}, and $10.9\%$ finding no difference.
A majority of turkers agreed $86.3\%$ of the time in the case of known characters, and $83.2\%$ for unknown.

\section{Analysis}

\subsection{Qualitative Inspection}

In Figure~\ref{fig:matrix} we show output from our model interpolating between a bold font and a light one, as well as a capital E and a $\Sigma$ simultaneously.
This demonstrates the smoothness of our manifolds and also suggests how they might offer support for font and character types not seen during training.
Figure~\ref{fig:qual} shows examples of reconstructions by models on two fonts for a variety of both known and unknown characters.
Our approach is more coherent and faithful than \textsc{EMD}, and \textsc{NN} is realistic but often stylistically incorrect.

\subsection{Chinese Radicals}

Figure~\ref{fig:tsne} shows t-SNE projections~\cite{tsne,bhtsne} of learned character embeddings colored by their radical, a sub-component of Chinese characters.  Radicals like \raisebox{-.3\height}{\includegraphics[trim=0 30 0 30, height=0.55cm]{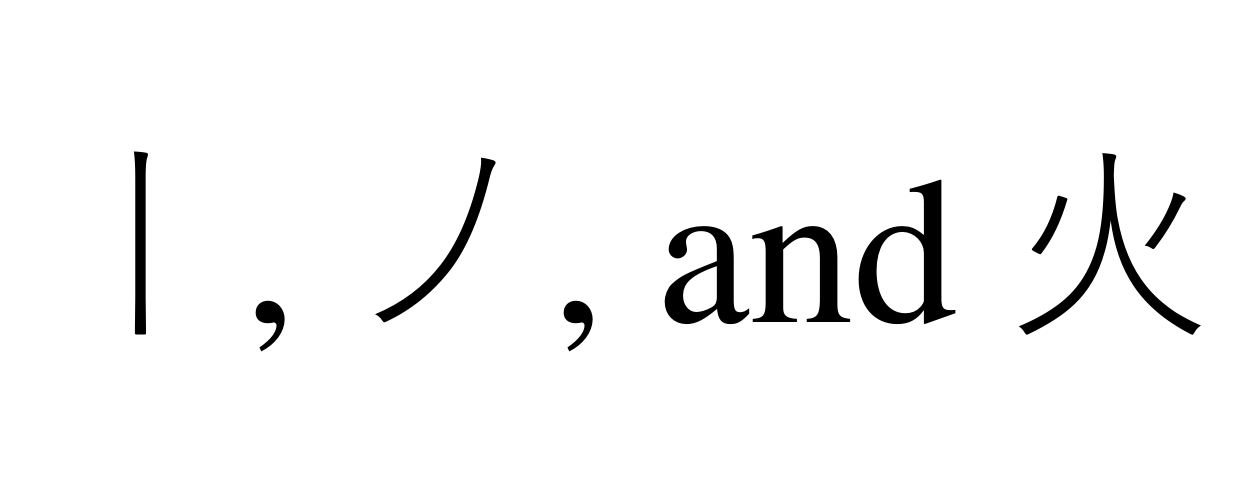}} --- which either share forms with others or can occur in different structures --- don't cluster together, while unique radicals like \raisebox{-.3\height}{\includegraphics[trim=0 30 0 30, height=0.55cm]{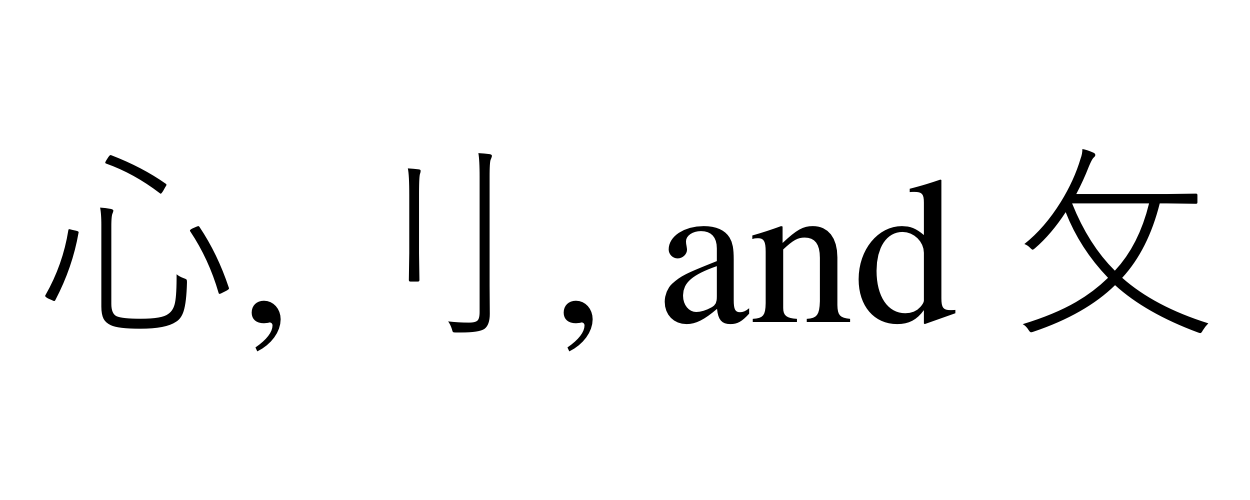}} do.

\section{Conclusion}
In this work introduced a generative model for typography capable of reconstructing characters in a novel font, of a novel shape, or both, and demonstrated its improvements over previous approaches on two datasets containing large numbers of characters.
We analyzed the results qualitatively, and inspected learned manifolds for smoothness.

In future work, this methodology has potential value not just to fonts, but to any domain which can also be factored over independent axes of variation, such as handwriting by different authors.
One could also incorporate this model into more complex downstream tasks such as OCR.
That being said, these domains also feature complex interactions between physically adjacent glyphs (our model treats different characters within a font as conditionally independent), so some further innovation would likely still be required.

There are also extensions to the model itself that might be worth exploring in future work, for instance operating on a stroke-based representation in order to perform reconstruction in the original TTF space instead of raw pixel space as we do here.
This would also likely assist with smoothness of edges and reduce the incidence of “corroded” output glyphs.

\section*{Broader Impact}


As our work can be used to augment or even replace the labor of human artists, it is worth discussing its potential broader impacts. The most obvious positive is that this technique can add value to font designers, by minimizing the overhead required to design a font that supports widespread internationalization. Our model's ability to interpolate stylistic properties can also make it easy to automatically generate completely novel fonts that are roughly similar to existing ones.

This also benefits speakers of languages that rely on less common glyphs, as it broadens their font selection. It can make it easier for them to both produce and consume digital content, allowing for better accessibility for demographics that currently have fewer options for orthographies they are most familiar with.

One potential negative impact is on the business of some font artists who cater to niche audiences that have less common glyph needs. Our model could potentially be used to replace such workers, and if so could also lead to less coherent renderings for uncommon orthographies if those who are not fluent in such scripts simply employ our system without a thorough understanding of the types of errors it may make.

\section*{Acknowledgements}
This project is funded in part by the NSF under grants 1618044 and 1936155, and by the NEH under grant HAA256044-17.

\bibliography{anthology,custom,acl2021}
\bibliographystyle{acl_natbib}

\clearpage
\appendix
\section{Architecture Details}
\label{app:arch}

\begin{table*}
\centering
\resizebox{\textwidth}{!}{%
\begin{tabular}{rcccccccc}
\toprule
  & \multicolumn{4}{c}{Google Fonts: Known Char} & \multicolumn{4}{c}{Google Fonts: Unknown Char} \\
\cmidrule(lr){2-5} \cmidrule(lr){6-9}
Observations  & 1 & 8 & 16 & 32 & 1 & 8 & 16 & 32  \\
\midrule
NN       & 405.15 & 258.11 & 227.91 & 207.04  & - & -  & - & -          \\
EMD      &  371.06 & 367.18  & 658.85  & 512.26 & 378.08 & 375.19 & 667.66 & 511.69 \\
Dual Manifold  & \textbf{275.56} & \textbf{202.58} &  \textbf{193.34} &  \textbf{189.94} & \textbf{276.47} & \textbf{212.76}  &  \textbf{205.34} & \textbf{202.91} \\
\midrule
  & \multicolumn{4}{c}{Chinese Simplified: Known Char} &  \multicolumn{4}{c}{Chinese Simplified: Unknown Char} \\
\cmidrule(lr){2-5} \cmidrule(lr){6-9}
NN       & 1086.58 & 908.58 & 883.18 &  872.52 & - & -  & - & -          \\
EMD      & 1013.80 & 1019.97  & 1288.32  & 1287.85 & 1303.96 & 1020.89 & 1303.92 & 1303.48 \\
Dual Manifold  & \textbf{916.41} & \textbf{879.03} &  \textbf{873.48} &  \textbf{868.41}  & \textbf{917.91} & \textbf{883.60}  &  \textbf{878.77} & \textbf{875.03} \\
\bottomrule
\end{tabular}
}
\caption{$L_2$ per glyph by number of observed characters for Google Fonts and Chinese Simplified.}
\label{tab:mse}
\end{table*}

The architectures of our encoder and decoder are largely identical to that of U-Net~\cite{unet} with key differences described here.
We find significantly improved results by inserting Instance Normalization layers~\cite{instancenorm} after convolution layers in our decoder.
We also replace the max pool layers within the encoder with blur pool layers~\cite{zhang2019making}.
As stated previously, we max pool the output of the encoders across character types or fonts, and then pass the flattened pooled representation through a fully connected layer to obtain the approximate posterior parameters $\mu$ and $\Sigma$.
A similar fully connected layer projects the character representation $Y_i$ to the appropriate size before being passed to the decoder.
As noted earlier, the parameters of the last two transposed convolutional layers in the decoder are dynamically output by MLPs which take as input the font representation $Z_j$.
These consist of a $256$ dimensional fully connected layer, a ReLU, and then a second fully connected layer to produce the relevant parameter.

We now provide further details on the specific layer sizes used in our model and inference network.
The following abbreviations are used to represent various components:
\begin{itemize}
    \item $F_i$ : fully connected layer with $i$ hidden units
    \item $R$ : ReLU activation
    \item $S$ : sigmoid activation
    \item $M$ : batch max pool
    \item $B$ : $2\times2$ spatial blur pool~\citep{zhang2019making}
    \item $C_i$ : convolutional layer with $i$ filters of $3\times3$, $1$ pixel zero-padding, stride of $1$
    \item $I$ : instance normalization
    \item $T_i$ : transpose convolution with $i$ filters of $2\times2$, stride of $2$
    \item $D_i$ : transpose convolution with $i$ filters of $2\times2$, stride of $2$, where kernel and bias are the output of an MLP (described below)
    \item $H$ : reshape to $-1\times256\times8\times8$
\end{itemize}

Our encoder is:

$C_{64}-R-C_{64}-R-C_{64}-R-B-C_{128}-R-C_{128}-R-B-C_{256}-R-C_{256}-R-B-C_{512}-R-C_{512}-R-B-M-F_{512}$

Our decoder is:

$F_{1024\times8\times8}-T_{1024}-C_{512}-I-R-C_{512}-I-R-T_{512}-C_{256}-I-R-C_{256}-I-R-D_{256}-C_{128}-I-R-C_{128}-I-R-D_{128}-C_{64}-I-R-C_{64}-I-R-C_1-S$

MLP to compute transpose convolutional parameter of size $j$ is:

$F_{256}-R-F_j$

\section{$L_2$ Results}
\label{app:mse}

In Table~\ref{tab:mse} we show results on Google Fonts and Chinese simplified for our model and baselines in terms of $L_2$.
Rankings are generally the same, and see that our approach performs best by this metric as well as SSIM.
We do however note that in places the $L_2$ numbers and SSIM numbers are not well correlated, and attribute this to $L_2$'s propensity for rewarding blurry output that minimizes expected distance over sharp output that may have slightly misaligned edges.

\end{document}